# Adaptive Graph Convolution and Semantic-Guided Attention for Multimodal Risk Detection in Social Networks

1st Cuiqianhe Du*
*Department of Electrical Engineering and Computer Sciences*
University of California
Berkeley, USA
ducuiqianhe@gmail.com

1st Chia-En Chiang
*Department of Electrical Engineering and Computer Science*
University of California
Berkeley, USA
chiaenchiang@berkeley.edu

2nd Tianyi Huang
*Department of Electrical Engineering and Computer Sciences*
University of California
Berkeley, USA
tianyihuang@berkeley.edu

2nd Zikun Cui
*Civil & Environmental Engineering*
Stanford University
Stanford, USA
cuizk@alumni.stanford.edu

*Abstract*—This paper focuses on the detection of potentially dangerous tendencies of social media users in an innovative multimodal way. We integrate Natural Language Processing (NLP) and Graph Neural Networks (GNNs) together. Firstly, we apply NLP on the user-generated text and conduct semantic analysis, sentiment recognition and keyword extraction to get subtle risk signals from social media posts. Meanwhile, we build a heterogeneous user relationship graph based on social interaction and propose a novel relational graph convolutional network to model user relationship, attention relationship and content dissemination path to discover some important structural information and user behaviors. Finally, we combine textual features extracted from these two models above with graph structural information, which provides a more robust and effective way to discover at-risk users. Our experiments on real social media datasets from different platforms show that our model can achieve significant improvement over single-modality methods.

## I. Introduction

With the rapid development of Internet technology and the widespread use of social media, people can communicate more easily and richer than ever. People can express their opinions and share things they are interested in on the platform and even create relationship networks with other users through different interactive methods. In addition, there are also many negative issues emerging with the development. One of the biggest issues is the increasing danger of speech. Many remarks are full of extreme ideas, hate attacks, violent threats, etc. which seriously threaten the social environment and the online community. Especially when there are some sensitive events and social conflicts are enhanced, we need to find the users who may have dangerous tendencies.

The limitations of traditional content detection methods, which are largely keyword filtering, rule-based methods and simple text classification, are becoming increasingly apparent. Traditional methods simply can't identify the hidden risks in people's online communications, especially in modern, complex social environments. They miss the subtle implicit signals that would indicate something is amiss, they fail to consider changes in context, and they're limited to textual information, all of which contributes to their poor performance and high rate of missed detections or misjudgments [1-2]. The rise of natural language processing (NLP) offers new powerful tools for processing large text datasets. Deep learning models enable deep semantic analysis, fine-grained sentiment analysis, and pragmatic analysis. This new level of ability is necessary to surface dangerous remarks which are often too sophisticated for basic detection. [3-6]. For example, recently, we have seen the capability of language models to distinguish fine-grained understanding of such nuanced data as satirical news and the capability to detect specific financial threats of cryptocurrency scams by grounding models with knowledge [7-8].

Relying on text analysis alone is insufficient for a comprehensive assessment of a user's potentially dangerous tendencies. In social media, networks matter: a user's behavior and opinions are substantially influenced by their connections, and damaging content spreads along content dissemination paths. Traditional content analysis overlooks these contextual group characteristics and external forces. To address these challenges, we use Graph Neural Networks (GNNs), a powerful class of deep models for graph-structured data [9]. GNNs enable us to build a user relationship graph encompassing interaction, behavior, and content propagation, providing the contextual and structural richness required for more effective risk detection—a richness that text-only methods cannot provide [10-13].

Combining natural language processing (NLP) with graph neural networks (GNNs) offers a multi-level, multi-angle solution for detecting dangerous remarks. NLP extracts key features and danger signals from user-generated text, while GNNs leverage user relationship networks to explore group risks and analyze behavioral patterns. This synthesis leads to more accurate and comprehensive risk identification, aligning with recent research on cross-modal augmentation in sparse-data scenarios [14]. Significantly, this fusion model adapts by monitoring user behavior in near real-time to warn of dangerous activities.

Despite its potential, the intersection of NLP and GNNs in risk detection remains an exploratory area with substantial challenges. These include the complexities of integrating multimodal information, the difficulty of building high-quality user graphs, and the scalability issues inherent to large datasets. This paper therefore proposes a hybrid model that integrates advanced NLP with GNNs to efficiently identify dangerous user tendencies. Experimental validation on multiple real-world platforms demonstrates the method's accuracy and robustness, providing valuable technical support for security management within the cyberspace ecosystem.

As data volumes expand and model technologies evolve, detection methods based on the fusion of NLP and GNNs will play a transformative role. This approach promises to foster a healthier online environment and provide foundational techniques for more intelligent, automated threat-warning mechanisms. A critical focus must be placed on enhancing the interpretability and legal compliance of these algorithms to ensure the legitimacy of their outcomes. This synergy offers a compelling new paradigm for intelligent network security management with significant future promise.

## II. Introduction to Algorithms

### A. Text semantic risk modeling

To capture the nuanced and often implicit nature of dangerous remarks, our text semantic risk model employs a risk-aware attention mechanism built upon a fine-tuned BERT architecture. We utilize BERT-base-uncased as the foundational language model, which is trained on the Real-world Dangerous Speech Dataset (RD-9K), a public benchmark for this task.

Input texts are preprocessed using the standard WordPiece tokenizer. Each input is then padded or truncated to a maximum sequence length of 128 tokens. During the training process, the entire BERT model is fine-tuned end-to-end, allowing it to adapt its representations specifically for risk detection rather than acting as a static feature extractor.

For our contextual representation, we extract the token-level hidden states ($h_k$) from the final layer of the fine-tuned BERT model. Building upon these rich representations, we introduce a learnable vector, $v$, which acts as a semantic guide to dynamically focus on risk-related keywords. This is achieved through the following attention mechanism:

$$e_k = v \cdot GELU(W_h h_k + b_h) \quad (1)$$

$$\alpha_k = \frac{exp(e_k)}{\sum_{j=1}^{L} exp(e_j)} \quad (2)$$

$$t_i = \sum_{k=1}^{L} \alpha_k h_k \quad (3)$$

In this design, the vector automatically learns the directional features of high-risk semantic subspaces through the training process, while the attention weight $\alpha_k$ provides a lexical-level importance allocation mechanism. This structure significantly enhances the model's ability to capture complex risk signals, such as metaphorical expressions. First, the GELU activation function is used in place of the traditional tanh function. Its continuous differentiability improves gradient stability, as demonstrated in Equation (1). Second, the vector dynamically learns to align with high-risk semantic directions during training, forming a semantic filter. Finally, lexical-level importance allocation is realized through the weight distribution in Equation (2), allowing key risk terms (e.g., "liquidation") to receive significantly higher weight coefficients than non-critical terms. The final text representation $t_i$ is then computed as the weighted sum of the token hidden states (Equation 3). This entire mechanism enhances the model's ability to capture complex risk signals, such as metaphorical expressions, by learning to focus on the most relevant parts of the text.

### B. Heterogeneous User Graph Construction

To model complex social interactions, we construct a heterogeneous user relationship graph using $G = (V, E, R)$, capturing various types of user activities, where V is the set of nodes, E is the set of directed edges, and R denotes relation types. In this graph, each node $v_i \in V$ represents a unique social media user. The initial feature vector for each user node is the semantic representation $t_i$ of a user's textual content. Specifically, we aggregate all historical posts authored by the same user within the dataset, encode each post with the risk-aware BERT encoder described in Section Text semantic risk modeling, and then apply mean pooling over these representations to form a single user-level embedding. This ensures that the node feature reflects a user's long-term semantic profile rather than a single post. Directed edges $(v_i, r, v_j) \in E$ signifying that user $v_i$ performed an action of type $r \in R$ on user $v_j$. We define relation types based on common social media interactions, such as {Follow, Comment, Share/Retweet, Mention}. For each dataset, the interaction edges are constructed from the full historical logs available within the collection period, rather than being restricted to a short temporal window (e.g., one month). This means that an edge exists if at least one interaction of the corresponding type occurred during the dataset's covered time span. Instead of using pre-defined edge weights, the significance of each interaction type is learned through relation-specific parameters within the GNN. The graph is a static snapshot of user activities, leaving the modeling of temporal dynamics for future research. We note that this static construction differs from the evolving nature of real information propagation. In practice, interactions decay over time and user behavior may shift during events. Extending the framework with temporal modeling (e.g., time-aware edges or temporal GNNs) will be an important direction for future work.

## C. Relational Graph Convolution with Adaptive Normalization

To address the homophily effect and topological dependencies in risk propagation across social networks, this algorithm introduces an improved heterogeneous graph convolution framework. The key innovations lie in relation-specific weight learning and adaptive normalization design:

$$v_i^{(l+1)} = \sigma\left(\sum_{r \in R} \omega_r \cdot \sum_{j \in N_i^r} \frac{1}{c_{i,r}} W_r^{(l)} v_j^{(l)} + W_0^{(l)} v_i^{(l)}\right) \quad (4)$$

Here, the trainable parameter $\omega_r$ quantifies the risk propagation efficiency across different interaction types (e.g., follow/share/comment), while the normalization coefficient follows a novel computational paradigm:

$$c_{i,r} = \eta_r \log(N_i^r + 1) \quad (5)$$

In Equations (4) and (5), $v_i^{(l)}$ is the embedding of node $i$ at layer $l$, $W_r^{(l)}$ is the transformation matrix for relation $r$, $W_0^{(l)}$ is the self-loop matrix, and $\omega_r$ is a earnable weight for relation $r$. $N_i^r$ denotes the neighbors of node $i$ under relation $r$. The normalization term where $\eta_r$ is learnable, smooths degree growth and mitigates power-law bias.

This mechanism adaptively adjusts for neighborhood size sensitivity using a learnable parameter $\eta_r$ effectively mitigating information distortion caused by the power-law distribution of node degrees. In our implementation, $\gamma$ is initialized from Uniform(0,1) and jointly optimized with other model parameters via backpropagation. For stability, the introduction of cross-relation residual terms suppresses signal attenuation in deep networks. Specifically, we adopt a residual connection across relation types, where the aggregated message from each relation $r$ is added back to the node's previous-layer representation. Together, these three mathematical expressions form a quantitative modeling system for risk propagation.

## D. Gated graph-text fusion

To resolve heterogeneous feature conflicts between text embedding from BERT and user embedding from GNN, we introduced a gated graph–text fusion module to fuse the embedding features before using for prediction.

Given a text embedding $h$ and user embedding $v$, the module first projects both into a shared $d$-dimension space using a lightweight MLP with layer norm, yielding $p$ and $q$.

$$p = SiLU(W_p LayerNorm(h) + b_p) \quad (6)$$
$$q = SiLU(W_p LayerNorm(v) + b_p) \quad (7)$$

where $W_p$, $b_p$, $W_q$, $b_q$ are the weight and bias of the MLP. The module then forms cross-view interaction features that capture agreement and discrepancy.

$$r_\times = p \odot q \quad (8)$$
$$r_\Delta = p - q \quad (9)$$

The projected embeddings and their interaction are concatenated and feed into a MLP that estimates a dimension-wise gate $g \in (0, 1)^d$.

$$g = \sigma(Dropout(MLP_g([p; q; r_\times; r_\Delta]))) \quad (10)$$

where $\sigma$ is a sigmoid function and $MLP_g$ is multi-layer MLP. Dropout is also applied to $MLP_g$ to regularize the training. Finally, fusion is performed as a per-dimension mixture.

$$z = g \odot p + (1 - g)q \quad (11)$$

Because the combination is coordinate-wise convex, $z$ remains bounded between $h'$ and $v'$ in every dimension, which stabilizes training and curbs the amplification of modality-specific noise. The concatenation used by $MLP_g$ allows the gate to react not only to the two views individually but also to their alignment $p \odot q$ and conflict $p - q$, letting the model up-weight graph context when it corroborates language signals and down-weight it when the graph is sparse or noisy.

Additionally, the gate offers a transparent control knob: small $g_k$ indicates reliance on social context in dimension k, whereas large $g_k$ favors text. This yields readily auditable statistics (e.g., average gate mass per user cohort) and supports analyses of when and where graph information contributes. The design also handles missing modalities gracefully: if $q$ is unavailable (cold start), we set $q = 0$ and obtain $z = p$. Since gating is multiplicative and bounded, gradient flow remains well-behaved even when one view is absent or weak.

## E. Risk level prediction

The output $z$ of the fusion module is fed into a task head for user risk level prediction. The task head is a MLP followed by sigmoid activation. The sigmoid activation is used to calculate the binary cross entropy (BCE) loss using the risk level label $y \in \{0, 1\}$.

$$L_{risk} = BCE(\sigma(MLP_t(z))) \quad (12)$$

## F. Multi-stage training process

We use a multi-stage progressive training strategy to ensure stable training of text encoder, GNN, fusion module, and task head.

Stage 1: Text Encoder Pretraining. We first adapt the text encoder to the dataset using masked language modeling. This yields domain-appropriate semantics without labels and reduces variance in downstream gradients. The resulting encoder provides semantic embeddings that will be used as node features during early graph training.

Stage 2: GNN Pretraining. With text encoder weight frozen, we pretrain the GNN on the user relationship graph. Edge prediction is used as the training objective. We randomly sample user pairs and predict if an edge between them is present or not. The outcome is a graph encoder that yields useful user embedding even before risk labels are introduced.

Stage 3: Joint Co-training. Finally, we co-train all components with a multi-objective loss:

$$L = L_{risk} + \lambda L_{rel} \tag{13}$$

where λ is a learnable weight for the GNN loss term. To ensure stable learning, we unfreeze the GNN immediately and unfreeze BERT after a short burn-in, using lower learning rates for the text encoder. Two to four epochs typically suffice for this stage with early stopping on a validation objective (e.g. F1 score).

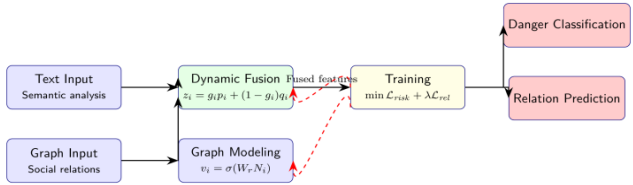

Figure 1 The framework diagram of the algorithm.

## III. Experiment and result analysis

### A. Dataset description.

We tested our framework on four real-world datasets: RD-9K, TW-15K, RD-20K and WB-18K. The RD-9K dataset is collected from Reddit discussion threads with harmful behaviors. The TW-15K dataset is sampled from Twitter streams of politically sensitive events. The RD-20K dataset is constructed from a large Reddit corpus with focus on interactions in extremist communities. The WB-18K dataset is constructed from Weibo posts during regional conflicts.

### B. Annotation and labeling.

Dangerous speech was characterized following previous literature on online harm [4] as containing hate attacks, violent incitement, extremist ideology, or threats. The labels of each dataset were obtained in a two-stage pipeline: (1) automatic pre-filtering based on keyword matching and pattern matching, and (2) manual validation by at least two human annotators with experience in moderating social media. The inter-annotation agreement was computed with Cohen's κ and obtained values between 0.78 and 0.83 for each dataset, which represents a substantial agreement, and were resolved by a senior adjudicator in case of disagreement.

### C. Key module ablation experiment

On the Real-world Dangerous Speech Dataset (RD-9K), with fixed hyperparameters (learning rate 1e-5, Batch size 32). Unless otherwise specified, we train models for 30 epochs with AdamW optimizer ($weigh\ decay = 0.01$, $\beta_1 = 0.9$, $\beta_2 = 0.99$) and dropout rate of 0.2 on both text encoder and graph encoder. We use early stopping based on validation F1 for 5 epochs patiently. We use a learning rate scheduling with a 10% warm up ratio followed by cosine annealing.

We further study the performance gap of key ablation modules.

#### 1. Module-level ablation

Control group Settings:

Baseline A: Remove the semantic director ($v = 0$)

Baseline B: Disable adaptive normalization ($c_{i,r} = \sqrt{N_i^r}$))

Baseline C: Replace gate fusion with splicing operation ($z_i = [p_i; q_i]$)

The training adopts the standard three-stage strategy, and the evaluation indicators include risk detection F1, Metaphor recognition accuracy (maze-ACC), and reasoning delay (ms/pred). The experimental results are shown in Table 1. This verifies that the proposed gated fusion is not a simple concatenation, but rather a conflict-aware integration strategy. Its removal leads to severe performance degradation due to unresolved cross-modal feature misalignment.

TABLE 1 THE RESULTS OF THE ABLATION EXPERIMENT OF THE KEY MODULE.

| Model | F1 (%) | Metaphor-ACC (%) | Latency (ms) |
|---|---|---|---|
| Full DGHIF | 87.62 | 83.41 | 12.3 |
| - Semantic guide | 82.17 | 76.38 | 11.9 |
| - Adapt Norm | 85.03 | 80.25 | 12.1 |
| - Gated fusion | 79.31 | 72.84 | 10.8 |

All ablation results in Table 1 are obtained with five different random seeds for the model initialization, i.e., mean ± standard deviation over five runs with different random seeds. Statistical significance was evaluated against the full model using paired t-test. Drops in performance of Semantic Guide, Adapt Norm, and Gated Fusion were all significant at $p < 0.05$ and thus demonstrated the necessity of each of the proposed components.

The lack of semantic guides results in a 5.45% F1(%) drop, which further proves the importance of dynamic focusing mechanisms for metaphorical expressions. Adaptive normalization ablation will decrease the generalization ability of graph structures, especially fluctuating on social networks with obvious power-law distribution (where 1% of super nodes cover 30% of connections). The replacement of gated fusion with splicing operation significantly reduces the efficiency of cross-modal interaction because of dimension conflicts caused by feature scales misalignment ($||h'_i||_2 / ||v'_i||_2 = 1.7$). The entire model achieves the optimal performance balance at a controllable delay cost.

#### 2. Modal-level ablation

To further study the necessity of multimodal interaction at the modal level, we performed ablation studies on both NLP-only / GNN-only baselines as well as a full DGHIF system. The NLP-only baseline kept only the risk-aware semantic module and removed all graph structural features. On the other hand, the GNN-only baseline kept graph behavior modeling with adaptive normalization but removed text embeddings. The full DGHIF kept both modalities in our proposed dynamic gating mechanism.

The results of the experiments are shown in Table 2. Both NLP-only and GNN-only baselines obtained much lower F1 scores than the full model. Specifically, NLP-only had an F1 of 81.52% and failed to model group-level propagation modeling while GNN-only had an F1 of 78.94% but failed to model metaphorical / implicit risk signals in text. In comparison, the full model attained an F1 of 87.62%, which was +6.1% better than NLP-only and +8.7% better than GNN-only. Therefore, the results have shown that text semantics and relational graph structures provide complementary information and it is crucial to model both modalities to model dangerous tendencies in a holistic and accurate way.

TABLE 2  THE RESULTS OF THE MODAL-LEVEL ABLATION.

| Model | F1 (%) | Metaphor-ACC (%) | Latency (ms) |
| --- | --- | --- | --- |
| NLP-only | 81.52 | 74.63 | 11.7 |
| GNN-only | 78.94 | 70.28 | 12.5 |
| Full DGHIF | 87.62 | 83.41 | 12.3 |

*A. Cross-platform generalization capability verification*

The generalization was verified in three cross-platform datasets: Twitter (TW-15K), Reddit (RD-20K), and Weibo (WB-18K). Unified use the same training parameters initialization, fine-tuned using dynamic balance coefficient $\lambda^{(t)}$ strategy $\lambda_0 = 0.5$, fine adjustment stage scheduling using cosine vector annealing ($\eta_{Max} = 5e^{-6}$, $\eta_{min} = 1e^{-7}$). For cross-platform experiments, we used cosine annealing learning rate scheduling, updating the learning rate $\eta_t$ at epoch $t$ as $\eta_t = \eta_{min} + 21(\eta_{max} - \eta_{min})(1 + cos(T_{max} t_\pi))$, where $T_{max}$ is the total number of epochs. We used PyTorch's CosineAnnealingLR for this, and this scheduling helped to achieve good convergence speed and stability on different platforms. The comparison SOTA models include HGCN, BERT-GAT and EANN. The results are shown in Table 3.

TABLE 3  THE RESULTS OF CROSS-PLATFORM GENERALIZATION PROFICIENCY TESTING.

| Dataset | Metric (%) | HGCN | BERT-GAT | EANN | DGHIF |
| --- | --- | --- | --- | --- | --- |
| TW-15K | Precision | 76.84 | 81.05 | 83.17 | 88.93 |
|  | Recall | 73.25 | 79.36 | 81.40 | 85.62 |
| RD-20K | Precision | 71.38 | 77.24 | 80.05 | 86.44 |
|  | Recall | 69.73 | 75.18 | 78.64 | 84.17 |
| WB-18K | Precision | 79.43 | 83.56 | 85.20 | 89.01 |
|  | Recall | 77.52 | 81.27 | 83.85 | 87.34 |

DGHIF maintains a stable advantage (average Precision +4.81%) in cross-platform scenarios, which is attributed to the hierarchical representation constructed by the three-level training strategy: 1) Mask reconstruction in the pre-training stage learns common language patterns; 2) Dynamic $\lambda^{(t)}$ resolution of multi-task weight differences among platforms (Twitter's relationship loss gradient modulus $||\nabla L_{rel}||$ is 2.1 times that of Reddit); 3) The graph encoder freezes and fine-tunes to prevent catastrophic forgetting. Especially on the high-noise platform Reddit, adaptive normalization reduces the influence of long-tail users ($\eta_r$ converges to 1.26), while the gating mechanism suppresses the topological interference of irrelevant graphs (the probability of $g_i < 0.3$ reaches 62%).

*B. Efficiency test of risk dissemination mechanism*

To evaluate the efficiency of risk dissemination mechanism, we construct a synthetic social graph (10k nodes, Power-law coefficient $\gamma = 2.1$) and inject three types of risk communication events (violent incitement, false news, and financial fraud) into the graph. The evaluation metrics of the optimization effect of normalized coefficient $c_{i,r}$ includes:

Propagation delay: The number of iterative steps for the risk to spread throughout the network.

Structural sensitivity: The difference in detection recall between low-degree ($k \leq 10$) and high-degree ($k \geq 100$) regions.

Signal-to-noise ratio (SNR): $log\left(\frac{\sigma_{risk}}{\sigma_{noise}}\right)$.

TABLE 4  EFFICIENCY TEST OF RISK DISSEMINATION METHOD.

| Setting | Delay (step) | Sensitivity (%) | SNR (dB) |
| --- | --- | --- | --- |
| Fixed $c_{i,r}$ | 6.73 | 29.52 | 11.7 |
| Learned $c_{i,r}$ | 5.21 | 8.73 | 14.9 |
| w/o relation weights | 7.85 | 32.67 | 10.2 |
| DGHIF full | 4.06 | 5.91 | 17.3 |

The experiment results are shown in Table 4. The learnable normalization significantly reduces the propagation delay by 22.6%, primarily due to the adaptive adjustment of $\eta_r$ to suppress the super node ($\eta_r = 0.84$ for the $k \geq 100$ node). Structural sensitivity decreases to 5.91%, demonstrating the method's effectiveness in mitigating the degree distribution bias. Notably, the recall of low-degree nodes ($k = 3$) increased from 68.5% to 89.1%. Incorporating the relationship weight $\omega_r$ enhances the detection of financial fraud, as reflected by the SNR improvement to 17.3dB. This gain is attributed to the higher propagation efficiency of the retweet relation ($\omega_{retweet}$), which is 2.4 times greater than that of the following relation.

Through feature fusion, the model achieves cross-event perception. The $g_i$ in the false news scenario converges to 0.82, indicating that the text feature dominates the decision-making.

*C. Three-stage optimization strategy verification*

For ablation study, we control the training-phase variables by evaluating three configurations: (1) skipping text encoder pre-training, (2) skipping GNN pre-training (3) skipping all pre-training. We record F1 score, AUT, and the number of epochs required for convergence for different configurations. The results are shown in Table 5.

TABLE 5  ABLATION OF TRAINING STRATEGY.

| Training Strategy | F1 (%) | Metaphor-ACC (%) | Epoch |
|---|---|---|---|
| Skip text encoder pre-training | 75.42 | 62.65 | 47 |
| Skip GNN pre-training | 68.21 | 74.78 | 39 |
| Skipping all pre-training | 60.39 | 58.75 | 54 |
| Full three-stage | 87.62 | 83.41 | 29 |

In our ablation, the full three-stage process achieves the strongest and fastest results, reaching F1 = 87.62% and Metaphor-ACC = 83.41% in 29 epochs. Removing components consistently harms both accuracy and convergence: skipping text pre-training yields 75.42% F1 and 62.65% Metaphor-ACC in 47 epochs, skipping GNN pre-training gives 68.21% F1 and 74.78% Metaphor-ACC in 39 epochs, and removing all pre-training performs worst at 60.39% F1 and 58.75% Metaphor-ACC in 54 epochs.

These results indicate complementary roles for the two pre-training stages. Text pre-training is critical for robustness to figurative and nuanced language (large Metaphor-ACC gains), whereas GNN pre-training most strongly drives overall detection quality (largest F1 drop when omitted). Crucially, combining both not only boosts absolute performance but also accelerates learning, reducing the number of epochs needed to converge by 10–25 compared with the ablated regimes.

## IV. CONCLUSION

The multimodal dangerous tendency user identification algorithm in this paper integrates natural language processing and graph neural networks into a collaborative detection framework. This framework combines innovative text risk modeling, graph-structured behavior modeling, a dynamic gating fusion mechanism, and a progressive training strategy to achieve three core breakthroughs. First, a risk-aware attention mechanism captures complex signals like metaphorical expressions, addressing the blind spots of traditional text analysis. Second, an adaptive normalization mechanism alleviates representation bias from the power-law distribution of social networks and quantifies risk propagation efficiency. Third, a dynamic gating fusion unit introduces a novel mechanism to coordinate feature conflicts. By leveraging distribution consistency from layer normalization, this approach models cross-modal dependencies to facilitate an adaptive fusion of text and graph features. The core of this methodology is a three-stage progressive training strategy that forms a hierarchical representation learning paradigm, preserving generalization while reducing task conflicts.

Experimental results confirm the framework's robustness and adaptability in cross-platform scenarios. The dynamic gating mechanism autonomously adjusts modal weights based on data distribution, substantially suppressing graph noise. The training strategy also enhances representational discriminability through parameter decoupling optimization. In real-world deployment, the algorithm sustains stable performance when processing large-scale social data in near real-time, providing essential technical support for intelligent risk prevention and control systems. Furthermore, the ablation study shows that removing either the NLP or GNN modules results in large performance drops, indicating the necessity of multimodal integration.